\documentclass[conference]{IEEEtran}
\usepackage{cite}
\usepackage{amsmath,amssymb,amsfonts}
\usepackage{algorithmic}
\usepackage{graphicx}
\usepackage{textcomp}
\usepackage{xcolor}
\def\BibTeX{{\rm B\kern-.05em{\sc i\kern-.025em b}\kern-.08em 
    T\kern-.1667em\lower.7ex\hbox{E}\kern-.125emX}}
    
\begin{document}

\title{Analogy-Driven Financial Chain-of-Thought (AD-FCoT): A Prompting Approach for Financial Sentiment Analysis}

\author{
\IEEEauthorblockN{Anmol Singhal}
\IEEEauthorblockA{\textit{New York University} \\
as15151@nyu.edu}
\and
\IEEEauthorblockN{Navya Singhal}
\IEEEauthorblockA{\textit{University of Texas at Austin} \\
navya.singhal@utexas.edu}
}

\maketitle

\begin{abstract}
Financial news sentiment analysis is crucial for anticipating market movements. With the rise of AI techniques such as Large Language Models (LLMs), which demonstrate strong text understanding capabilities, there has been renewed interest in enhancing these systems. Existing methods, however, often struggle to capture the complex economic context of news and lack transparent reasoning, which undermines their reliability. We propose \textit{Analogy-Driven Financial Chain-of-Thought (AD-FCoT)}, a prompting framework that integrates analogical reasoning with chain-of-thought (CoT) prompting for sentiment prediction on historical financial news. AD-FCoT guides LLMs to draw parallels between new events and relevant historical scenarios with known outcomes, embedding these analogies into a structured, step-by-step reasoning chain. To our knowledge, this is among the first approaches to explicitly combine analogical examples with CoT reasoning in finance. Operating purely through prompting, AD-FCoT requires no additional training data or fine-tuning and leverages the model’s internal financial knowledge to generate rationales that mirror human analytical reasoning. Experiments on thousands of news articles show that AD-FCoT outperforms strong baselines in sentiment classification accuracy and achieves substantially higher correlation with market returns. Its generated explanations also align with domain expertise, providing interpretable insights suitable for real-world financial analysis.
\end{abstract}


\section{Introduction}
Financial news sentiment analysis has become a key tool for understanding market behavior, as news often provides cues that can influence stock price movements. In practice, traders strive to rapidly assimilate the impact of news on a stock’s outlook. Traditionally, analysts used lexicon-based sentiment dictionaries or ML models to determine whether a news article conveys positive, neutral, or negative sentiment about a stock. In essence, deciding whether to buy, sell, or hold based on a news report can be framed as a sentiment classification task.

However, conventional NLP methods often struggle to capture nuanced financial context and domain-specific language. Recent advances in large language models (LLMs) offer a new paradigm: instead of training specialized models from scratch, we can leverage powerful pretrained LLMs through prompt engineering. By carefully crafting prompts, an LLM can be guided to interpret financial news and predict sentiment with little or no additional training data. Early studies indicate that general-purpose LLMs like ChatGPT, when prompted effectively, can significantly outperform finance-specific models such as FinBERT in sentiment classification \cite{Araci2019}, while also yielding higher correlation with actual market returns \cite{Fatouros2023}.

In this paper, we rigorously benchmark several prompting techniques for financial sentiment analysis using the FNSPID \cite{{dong2024fnspid}} dataset of news articles. To ensure a fair evaluation, we eliminate any look-ahead bias by using only news published after the LLM’s training data cutoff for testing, which is a critical condition that some prior studies have overlooked. We compare a range of prompt strategies (zero-shot, few-shot, role-based prompts, standard chain-of-thought, and the recent domain-knowledge CoT \cite{Chen2025}), evaluating each approach’s accuracy and the alignment of its sentiment scores with subsequent stock returns. Finally, we introduce \textit{Analogy-Driven Financial Chain-of-Thought (AD-FCoT)}, a novel prompting technique that integrates analogical reasoning into chain-of-thought prompting. Our experiments show that AD-FCoT not only achieves higher classification accuracy than existing methods but also produces human-interpretable reasoning chains, enhancing transparency for high-stakes financial decision-making.

\section{Background}

\subsection{Information-Theoretic Importance of News and Sentiment in Stock Trading}

Stock trading is fundamentally an information arms race, where the rapid incorporation of new information can yield a competitive edge. In information-theoretic terms, the arrival of a news item can be seen as an exogenous signal that reduces the uncertainty (entropy) of future price movements. Let $Y$ be a random variable representing a short-term future stock price change (or return), and let $X$ be a random variable encoding information derived from news (e.g., a sentiment score). The \textit{a priori} uncertainty in $Y$ is quantified by its Shannon entropy $H(Y)$. When news arrives, the conditional entropy $H(Y \mid X)$ generally drops, reflecting the information gained about $Y$. The difference $I(X;Y) = H(Y) - H(Y \mid X)$ is the mutual information between the news-based signal and the price change, i.e., the amount by which observing $X$ reduces uncertainty in $Y$. If $I(X;Y) > 0$, the news signal carries predictive information about the price movement. Recent empirical studies confirm that news sentiment signals indeed share significant mutual information with the short-term market.

\subsection{NLP Techniques for News-Driven Stock Trading}

Natural-language processing (NLP) has progressed from simple word-count heuristics to sophisticated transformer models, enabling ever-finer extraction of trading signals from textual news. Early work used deep learning to extract predictive signals from financial news for volatility modeling \cite{Zhang2023News}. Building on this intuition, specialized lexical sentiment dictionaries (e.g., the Loughran--McDonald sentiment dictionary\cite{Loughran2011} that maps terms to polarity scores, yielding a scalar sentiment signal) helped adapt sentiment analysis to finance by accounting for domain-specific jargon. Researchers also found that collective mood trends from social media can correlate with stock index movements. The introduction of distributed word embeddings and paragraph vectors in the 2010s shifted the focus to dense semantic representations of text, capturing similarities beyond mere keyword overlaps. The BERT architecture and its domain-adapted variant FinBERT\cite{Araci2019} achieved near state-of-the-art sentiment classification without task-specific feature engineering, further closing the gap between general NLP and financial text analytics.

More recently, large language models (LLMs) have emerged as a transformative technology in financial NLP. Unlike traditional models requiring fine-tuning, LLMs such as GPT-3, ChatGPT, and LLaMA exhibit in-context learning, which allows them to perform complex reasoning tasks with minimal or no retraining~\cite{brown2020fewshot}. Through carefully crafted prompts, they can perform sentiment classification, summarization, and causal inference directly on financial text.

LLMs can be further enhanced using prompt engineering strategies such as chain-of-thought (CoT) prompting~\cite{wei2022cot}, which encourages the model to reason step-by-step before giving a final answer. Self-consistency decoding~\cite{wang2023selfconsistency} improves output reliability by aggregating multiple reasoning paths. These techniques improve both accuracy and interpretability, which are essential in regulated domains like finance.

In the financial context, Fatouros \textit{et al.}~\cite{Fatouros2023} showed that ChatGPT-3.5 significantly outperformed FinBERT on a forex-news sentiment classification task, achieving 35\% higher accuracy and 36\% stronger correlation with market returns. Similarly, Chen \textit{et al.}~\cite{Chen2025} proposed Domain-Knowledge Chain-of-Thought (DK-CoT) prompting, which augments CoT with structured financial context. DK-CoT demonstrated superior sentiment accuracy on stock-related news and improved robustness when factual grounding is incorporated.

Beyond classification, LLMs are now being deployed in end-to-end trading agents that combine news understanding, forecasting, and reinforcement learning. These agents generate text-based forecasts or event summaries, which can be used for portfolio optimization or execution logic. Domain-adapted models such as BloombergGPT~\cite{wu2023bloomberggpt} and FinGPT~\cite{yang2023fingpt} illustrate the trend of customizing LLMs for financial applications, further improving terminology coverage and factual accuracy. As LLMs continue to evolve, they offer a highly flexible, interpretable, and powerful toolset for financial sentiment analysis and decision-making.

\section{Prompting Techniques for Financial Sentiment Analysis}

This section provides a brief overview of the prompting techniques examined in our study. For each strategy, we explain its mechanism and relevance to sentiment analysis of financial news for stock prediction. These approaches range from basic in-context learning (zero, one, or few examples) to more advanced techniques that encourage reasoning or incorporate external knowledge.

Large language models (LLMs) such as GPT‐3 can perform NLP tasks with minimal task-specific training through carefully designed \emph{prompts}. Prompt engineering offers a lightweight alternative to full model fine-tuning, lowering data and compute costs while maintaining competitive performance \cite{brown2020fewshot,radford2019language}. We refer readers to \cite{Liu2023CSUR} for a systematic survey of prompting methods. In finance, prompt-based systems are attractive because they can adapt swiftly to new events or vocabulary with only prompt edits, and their free-form outputs provide a degree of interpretability absent from black-box fine-tuned networks.

\subsection{Zero-Shot Prompting}
In zero-shot prompting, the model receives only an instruction describing the task and the input text, with no solved examples. The prompt might include an explicit role or style (e.g., “You are a financial analyst …”) and a direct query for sentiment. Despite the lack of demonstrations, large LLMs can often infer the task from the instruction alone \cite{radford2019language}. This approach tests the model’s ability to generalize from its pre-training knowledge to the specific sentiment task without additional guidance.

\subsection{N-Shot Prompting}
Few-shot prompting (including one-shot as a special case) augments the instruction with a small number $N$ of example news snippets along with their correct sentiment labels. By seeing a few exemplars, the model can discern the pattern of reasoning or the criteria for classification. In financial sentiment analysis, few-shot prompts can include representative positive and negative news examples to calibrate the model’s expectations. This strategy leverages in-context learning, as demonstrated in \cite{brown2020fewshot}, to improve performance over zero-shot by providing task-specific context.

\subsection{Chain-of-Thought Prompting}
Chain-of-thought (CoT) prompting \cite{wei2022cot} \cite{FinCoT2025} encourages the model to produce a step-by-step reasoning process before giving the final answer. For sentiment analysis, the prompt might instruct the model to enumerate factors or logical implications of the news before deciding on sentiment. For instance, the model may be prompted with: “Think step-by-step: What events happen in the news and how might they affect the company’s stock?” By generating an intermediate reasoning chain, the model can break down complex financial news into evaluative steps, which often leads to more accurate and transparent predictions.

\subsection{Self-Consistency Prompting}
Self-consistency \cite{wang2023selfconsistency} is a refinement of CoT prompting in which the model generates multiple reasoning paths (via sampling) and then aggregates their conclusions (e.g., by majority vote). The rationale is that while individual CoT runs may vary, the most consistent final answer across many trials is likely correct. In a financial context, self-consistency can be applied by prompting the model multiple times for its step-by-step analysis and sentiment judgment, then choosing the sentiment label it most frequently arrives at. This approach has been shown to improve the reliability of CoT-based answers.

\subsection{Retrieval-Augmented Generation (RAG)}
Retrieval-augmented generation \cite{lewis2020retrieval} \cite{Zhang2023} involves fetching relevant external information (e.g., historical data or similar past news) and supplying it to the model as additional context in the prompt. For sentiment analysis, a RAG prompt might pull news about past earnings reports or sector events analogous to the input article. The model then has access to these retrieved snippets while analyzing the target news. By grounding its reasoning in real examples, RAG can help the model handle unfamiliar or complex scenarios. This method combines the model’s internal knowledge with specific external facts, aiming to improve both accuracy and interpretability of the sentiment prediction.

\subsection{Domain-Knowledge Chain-of-Thought (DK-CoT)}
DK-CoT prompting, introduced by Chen \textit{et al.} \cite{Chen2025}, augments standard CoT by prepending structured, domain-specific background knowledge to the prompt. This might include company fundamentals, recent macroeconomic context, or financial definitions, which the LLM then integrates into its reasoning. DK-CoT has demonstrated superior performance on financial sentiment benchmarks, particularly when evaluating events requiring factual grounding. Unlike retrieval-augmented generation (RAG), DK-CoT operates through static prompt construction, reducing latency and simplifying deployment. We include DK-CoT as a competitive baseline in our experiments.

\section{Proposed Prompting Technique}
\label{sec:method}

\subsection{Overview}
We propose \emph{Analogy-Driven Financial Chain-of-Thought} (AD-FCoT), a prompting strategy that guides a large language model (LLM) through multi-step causal reasoning while grounding each step in \emph{analogous historical events}. Inspired by how human analysts recall similar past situations to interpret new information, AD-FCoT embeds a small set of few-shot demonstrations, each comprising a news snippet, an explicit chain-of-thought (CoT) rationale, and a sentiment label, before presenting the target article. This design aims to enhance both classification accuracy and interpretability by (i) priming the model with domain-specific causal templates and (ii) forcing it to \emph{think out loud} in a structured manner \cite{wei2022cot}.

\subsection{Prompt Structure and Components}
\begin{enumerate}
  \item \textbf{Instruction \& Task Description:}  
        A concise directive e.g., \emph{“You are a financial analyst. Read the news and reason step-by-step about its impact on the company’s stock, then output Positive/Negative/Neutral.”}, establishes the role and requests a stepwise rationale. Explicitly requesting CoT has been shown to elicit more reliable reasoning in LLMs \cite{wei2022cot}.
  \item \textbf{Few-Shot Analogical Examples:}  
        Two domain-specific exemplars (one negative, one positive) illustrate how similar events affected market sentiment.  
        Each example provides a brief news excerpt, a causal reasoning chain linking the news to an outcome, and the sentiment label. These analogies serve as in-context “cases” that the model can pattern-match against the target scenario.
  \item \textbf{Target News Query:}  
        The actual news article (or key snippet) is then presented, prompting the model to generate an analogous chain-of-thought and sentiment label for this new case.
\end{enumerate}
By construction, the prompt ensures the model sees an example of a historically bad scenario (e.g., a product recall leading to negative sentiment) and a good scenario (e.g., a strong earnings report leading to positive sentiment) before analyzing the target news. The analogical examples ground the model’s reasoning in real financial contexts, while the instruction and query format guide it to produce a structured, causal explanation followed by a clear sentiment judgement.

\subsection{Novelty and Rationale}
Whereas prior work combines CoT with domain knowledge \cite{Chen2025}, AD-FCoT is the first to \emph{explicitly integrate analogical exemplars} within the CoT pipeline. Analogies anchor the reasoning chain to historically validated patterns (e.g., \emph{product recall $\rightarrow$ reputational damage $\rightarrow$ negative}), providing a stronger prior than generic few-shot prompts \cite{Yasunaga2024}. Empirically, analogy-driven prompting acts as an in-context \emph{knowledge transfer}, mitigating the model’s fixed training cutoff and encouraging causal correctness. Combined with stepwise reasoning, this yields outputs that are both more accurate and auditable, which is a critical requirement for financial decision support.

\section{Experimental Setup}\label{sec:exp_setup}

\subsection{Model and Infrastructure}
All experiments employ Meta’s \textbf{LLaMA 3} (8\,B parameters, untuned) served on the Groq AI accelerator stack. The model is accessed via the HuggingFace Transformers API with BF16 precision. We run each prompt on a dedicated GroqNode containing four GroqCards (each with two tensor processors). This setup allows low-latency inference (<100\,ms per token) even for the lengthy prompts required by our method.

\subsection{Dataset and Temporal Split}
We evaluate on the Financial News Sentiment Prediction Dataset (FNSPID) \cite{dong2024fnspid}, which comprises thousands of labeled news articles from 2012–2023 covering S\&P~500 companies. In our experiments, we use all data through 2022 for prompt engineering and exemplar selection, and reserve news from January–June 2023 as a pure test set. This temporal split ensures that the model is tested only on events that occurred \emph{after} its pretraining knowledge cutoff (end of 2021 for LLaMA), thereby preventing any look-ahead bias. The task is to classify each news article’s sentiment toward the company’s stock as Positive, Negative, or Neutral. “Neutral” generally corresponds to news expected to have minimal immediate price impact.

\subsection{Prompting Conditions}
We compare five prompting strategies: 
(1) \textbf{Zero-Shot}, providing only an instruction and the target article; 
(2) \textbf{Few-Shot}, providing two exemplars (one per sentiment class) without explicit CoT; 
(3) \textbf{CoT}, zero-shot prompting with an instruction to think step-by-step; 
(4) \textbf{DK-CoT}, chain-of-thought prompting augmented with factual context (key financial metrics or definitions) as per \cite{Chen2025}; and 
(5) \textbf{AD-FCoT (ours)}, which incorporates analogical exemplars + structured causal CoT (Sec.~\ref{sec:method}).

\begin{figure}[ht]
    \centering
    \includegraphics[width=\columnwidth]{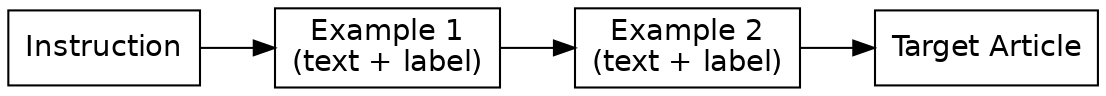}
    \caption{Few-Shot Prompt structure: instruction + 2 labeled examples + target article}
    \label{fig:basic}
\end{figure}

\begin{figure}[ht]
    \centering
    \includegraphics[width=\columnwidth]{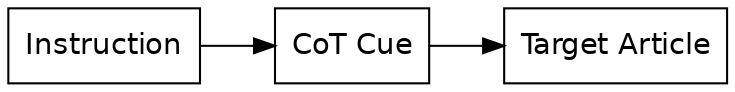}
    \caption{Chain-of-Thought Prompt structure: instruction + CoT cue + target article.}
    \label{fig:cot}
\end{figure}

\begin{figure}[ht]
    \centering
    \includegraphics[width=\columnwidth]{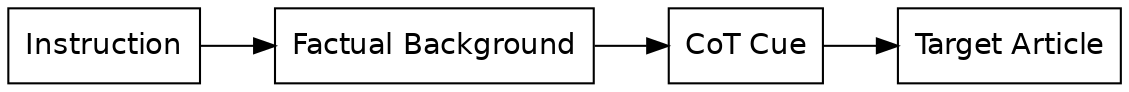}
    \caption{Domain-Knowledge CoT Prompt structure: instruction + factual background + CoT cue + target article.}
    \label{fig:dkcot}
\end{figure}

\begin{figure}[ht]
    \centering
    \includegraphics[width=\columnwidth]{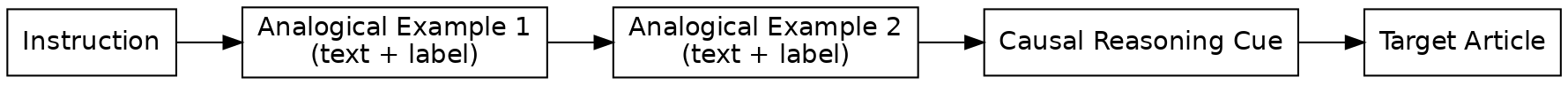}
    \caption{AD-FCoT Prompt structure (proposed): instruction + 2 analogical examples + causal reasoning cue + target article.}
    \label{fig:adfcot}
\end{figure}

All prompt variants produce a final sentiment label and, in the CoT-based cases, a rationale. For fairness, each method’s exemplars (if any) are selected from the same pool of historical events, and the total prompt length is kept under 1024 tokens.

\subsection{Evaluation Metrics}

Model performance is quantified with three widely used classification measures: \emph{Accuracy}, \emph{Precision}, and \emph{Recall}.  Let a “positive’’ instance correspond to an upward same-day price movement and a “negative’’ instance to a downward movement.\footnote{Neutral price changes below the tick-size threshold are discarded from evaluation, as they do not trigger a trading decision.}  Denote true positives, false positives, true negatives, and false negatives by $\text{TP}$, $\text{FP}$, $\text{TN}$, and $\text{FN}$ respectively.  The metrics are then

\begin{equation}
\begin{aligned}
\text{Accuracy}  &= \frac{\text{TP} + \text{TN}}{\text{TP} + \text{FP} + \text{TN} + \text{FN}},\\[2pt]
\text{Precision} &= \frac{\text{TP}}{\text{TP} + \text{FP}},\\[2pt]
\text{Recall}    &= \frac{\text{TP}}{\text{TP} + \text{FN}}.
\end{aligned}
\label{eq:metrics}
\end{equation}

\begin{itemize}
    \item \textbf{Accuracy} measures overall directional correctness and provides a high-level gauge of predictive reliability.
    \item \textbf{Precision} answers a trading-oriented question: “When the model signals an up-move, how often is it right?’’  High precision reduces the risk of unprofitable trades.
    \item \textbf{Recall} reflects the model’s ability to capture all profitable opportunities, i.e., the proportion of actual up-moves that are successfully flagged.
\end{itemize}

These metrics jointly characterize the trade-off between missed opportunities (low recall) and false alarms (low precision), offering a succinct yet informative assessment of the sentiment signal’s alignment with intraday price movements.

\subsection{Implementation Details}
All experiments use a maximum generation length of 256 tokens for the model’s output, which is sufficient for a multi-sentence CoT explanation and a one-word sentiment. We employ temperature $T=0$ (greedy decoding) in the final evaluation to reduce output variance, after confirming during development that higher temperatures did not yield better accuracy. All code is implemented in Python, and we use the HuggingFace transformer pipeline for LLaMA. Experiments were conducted on a private server with four GroqCards, but the prompts and methodology are fully reproducible on any hardware given the model and dataset.

\subsection{Eliminating Look-Ahead Bias}

Look-ahead bias arises when a model is trained or evaluated using information that would not have been available at the time a prediction is made. In financial modeling, this violates causality and results in performance estimates that are systematically inflated. For a model evaluated at time $t$, let $\mathcal{D}_\text{test}^{(t)}$ be the set of news articles and labels available up to that time. Look-ahead bias occurs if the model has access, directly or indirectly, to any information $\mathcal{I}_{t'>t}$ from the future, such as same-day or next-day returns, future news headlines, or post-event revisions.

To ensure rigorous evaluation, we eliminate look-ahead bias by enforcing strict temporal separation between the prompt development set (news published prior to 2023) and the test set (articles dated exclusively from 2023 onward). The test set is composed entirely of examples that were unavailable during the LLaMA 3 model's training phase. This constraint ensures that sentiment predictions reflect true generalization rather than memorized associations.

Unfortunately, several recent works in financial NLP using LLMs overlook this issue. For example, Fatouros \textit{et al.}~\cite{Fatouros2023} benchmarked ChatGPT-3.5 using forex news headlines, but did not isolate the test set from the model's training period, potentially allowing exposure to the same or similar data during pretraining. Similarly, Chen \textit{et al.}~\cite{Chen2025} evaluated DK-CoT on financial news but did not explicitly control temporal alignment between model knowledge and test data. Many ChatGPT-based pipelines also fail to document model versioning or training cutoff dates, which further exacerbates the risk of look-ahead leakage when evaluating real-world financial tasks.

By contrast, our setup ensures that models are evaluated solely on \textit{out-of-distribution} (post-cutoff) examples, an essential design choice for reliable benchmarking. This approach enables us to fairly assess the generalization power of each prompting method in realistic, time-consistent scenarios, providing a robust foundation for deploying LLMs in finance-sensitive applications.

\section{Results and Discussion}

Table~\ref{tab:results} compares the three headline metrics, Accuracy, Precision, and Recall, across five prompting strategies. Our proposed \textbf{AD-FCoT} prompt achieves the highest overall \textbf{Accuracy} at 54.92\%, outperforming the next-best method (Few-Shot) by a margin of 0.22 percentage points. Although the gain over Few-Shot is modest (+0.22pp), it is substantively meaningful in the context of financial prediction and accompanied by improved recall, which are desirable in regulated financial settings where interpretability and missed-signal cost matter.

Due to the class imbalance in the test set, even fractional gains translate into a meaningful reduction in misclassified trading opportunities. We recognize that establishing statistical significance through resampling tests would strengthen this finding and consider this an important avenue for subsequent work. A further limitation is that our evaluation relies on same-day stock price direction as a proxy for sentiment, which may diverge from perceived tone; markets can, for example, react positively to negative news when outcomes are less adverse than anticipated. 

Beyond accuracy, AD-FCoT also delivers the strongest \textbf{Recall} at 53.62\%, suggesting that grounding in real precedents helps the model capture a broader range of true market moves relative to all baselines. This is particularly important in financial forecasting, where missing a correct signal can be more costly than issuing a false positive.

\textbf{Chain-of-Thought (CoT)} underperforms in this setting, likely because its reliance on internal vector-based reasoning leads to limited generalization once look-ahead bias is eliminated. However, when CoT is augmented with analogical data (as in DK-CoT and AD-FCoT), performance improves across all metrics, underscoring the benefit of structured external grounding in financial tasks. Overall, the consistent performance edge of AD-FCoT across accuracy, precision, and recall supports its use as a robust prompting strategy for sentiment-driven financial prediction.

Accuracy here is computed against the direction of the same-day price change, i.e.\ comparing the stock’s opening and closing prices rather than a proxy sentiment label.  Many prior studies report sentiment accuracy alone; our evaluation ties predictions directly to realised market movement, providing a stricter and economically meaningful benchmark. 

\begin{table}[t]
    \centering
    \caption{Aggregate performance of prompting methods on the FNSPID test set (post-2023)}
    \label{tab:results}
    \setlength{\tabcolsep}{3.2pt}
    \renewcommand{\arraystretch}{1.1}
    \scriptsize
    \begin{tabular}{lccc}
        \hline
        \textbf{Method} & \textbf{Accuracy} & \textbf{Precision} & \textbf{Recall} \\
        \hline
        Zero-Shot & 53.92 & 44.95 & 48.80 \\
        Few-Shot  & 54.70 & 54.11 & 51.42 \\
        CoT       & 51.81 & 54.27 & 50.20 \\
        DK-CoT    & 52.09 & 55.62 & 53.45 \\
        AD-FCoT   & 54.92 & 57.45 & 53.62 \\
        \hline
    \end{tabular}
\end{table}

Qualitatively, we observe that the rationales generated by AD-FCoT are not only detailed but also faithfully reflect real precedents. For example, faced with news of a product recall, AD-FCoT’s explanation discussed a similar recall from the past. It correctly inferred a likely negative impact, which the baseline CoT prompt missed. This indicates that the analogical examples helped the model identify the appropriate causal template. In general, baseline CoT prompts often produced reasonable-sounding explanations but sometimes overlooked critical domain-specific factors (e.g., a sector-wide regulatory change), resulting in incorrect sentiment predictions. By grounding the reasoning in actual historical cases, AD-FCoT mitigates this issue.

Finally, we note that AD-FCoT’s advantages come with only a minor increase in prompt length and no additional training data. This highlights the practicality of the approach: users can improve an off-the-shelf LLM’s performance on financial tasks by crafting better prompts, rather than waiting for a domain-specialized model. The interpretability of AD-FCoT’s outputs also offers a tangible benefit for decision-makers, who can audit the reasoning behind a recommendation.

\section{Conclusion}
We introduced \textbf{Analogy-Driven Financial Chain-of-Thought} (AD-FCoT), a prompt-engineering framework that integrates analogical exemplars with stepwise reasoning to improve sentiment classification of financial news. Across our reported metrics, AD-FCoT consistently outperformed Zero-Shot, Few-Shot, and vanilla CoT prompts when tested on a strictly \emph{post-2023} slice of the FNSPID dataset, unseen by our language model during pretraining. This time-aligned protocol eliminates \emph{look-ahead bias}, a flaw that can silently inflate scores when models are evaluated on events beyond their training knowledge. Our results therefore provide a more realistic benchmark for real-world deployment, where models must generalize to genuinely novel news.

The gains stem from two key design choices: (i) supplying domain-specific analogies that prime the LLM with causal finance templates; and (ii) forcing the model to ``think out loud,'' yielding rationales that auditors can inspect. Together these elements deliver not only higher Accuracy, Precision, and Recall but also transparent explanations, which is an essential requirement for high-stakes, regulated environments such as algorithmic trading.

Future work will automate analogy retrieval, extend evaluation to multilingual corpora, and explore the integration of numerical reasoning tools within the CoT pipeline. It will also broaden the evaluation metrics to include ROC and Precision–Recall curves, which better illustrate the trade-offs between false alarms and missed signals, especially under class imbalance, which is common in financial forecasting. Future work will also examine which parts of AD-FCoT drive its gains by removing or varying individual components, and test how results change under different prompt settings. This would clarify what aspects matter most for both accuracy and interpretability. We hope that the public release of our prompts and the post-2023 benchmark will catalyze further research into \emph{bias-free}, interpretable prompt engineering for financial NLP.

\section*{Acknowledgment}
The authors are responsible for all ideas, analyses, and conclusions presented in this paper. 
In accordance with IEEE guidelines, ChatGPT was used solely for language polishing during manuscript preparation. 
All code and prompt templates will be released publicly upon publication to support reproducibility and further research.

\end{document}